\documentclass[11pt,a4paper]{article}
\usepackage[hyperref]{naaclhlt2019}
\usepackage{times}
\usepackage{latexsym}

\usepackage{pdfsync}
\usepackage{amsmath}
\usepackage{amssymb}
\usepackage{graphicx}
\usepackage{microtype}
\usepackage{url}
\usepackage{xspace}
\usepackage{todonotes}
\usepackage{booktabs}
\usepackage{subcaption}
\usepackage{multirow}

\aclfinalcopy 

\newcommand{\fscore}{F$_1$-score\xspace}

\newcommand{\savingspace}[1]{}
\title{What do Entity-Centric Models Learn? Insights from Entity
  Linking in Multi-Party Dialogue}

\author{
 	Laura Aina\thanks{~~denotes equal contribution.}\hspace{4ex}Carina Silberer\footnotemark[1]\hspace{4ex}Ionut-Teodor Sorodoc\footnotemark[1]\vspace{1ex}\\ 
 	 \hspace{3ex}\textbf{Matthijs Westera\footnotemark[1]} \hspace{3ex}\textbf{Gemma Boleda}\vspace{2ex} \\
 	Universitat Pompeu Fabra \\
 	Barcelona, Spain \\
 	{\tt \{firstname.lastname\}@upf.edu}	
 }

\date{}

\begin{document}
\maketitle

\begin{abstract}
Humans use language to refer to entities in the external world. 
Motivated by this, in recent years several models that incorporate a bias towards learning entity representations have been proposed.
Such entity-centric models have shown empirical success, but we still know little about why.

In this paper we analyze the behavior of two recently proposed entity-centric models in a referential task, Entity Linking in Multi-party Dialogue (SemEval 2018 Task 4).
We show that these models outperform the state of the art on this task, and that they do better on lower frequency entities than a counterpart model that is not entity-centric, with the same model size.
We argue that making models entity-centric naturally fosters good architectural decisions.
However, we also show that these models do not really build entity representations and that they make poor use of linguistic context. 
These negative results underscore the need for model analysis, to test whether the motivations for particular architectures are borne out in how models behave when deployed.
\end{abstract}

\section{Introduction}

Modeling reference to entities is arguably crucial for language understanding, as humans use language to talk about things in the world.
A hypothesis in recent work on referential tasks such as co-reference resolution and entity linking~\cite{haghighi2010coreference,clark2016improving,henaff16:entitynetworks,aina2018semeval,clark2018neural} is that encouraging models to learn and use entity representations will help them better carry out referential tasks.
To illustrate, creating an entity representation with the relevant information upon reading \emph{a woman} should make it easier to resolve a pronoun mention like \textit{she}.%
\footnote{Note the analogy with traditional models in formal linguistics like Discourse Representation Theory \cite{kamp2013discourse}.}
In the mentioned work, several models have been proposed that incorporate an explicit bias towards entity representations.
Such \textbf{entity-centric} models have shown empirical success, but we still know little about what it is that they effectively learn to model.
In this analysis paper, we adapt two previous entity-centric models \cite{henaff16:entitynetworks,aina2018semeval} for a recently proposed referential task and
show that, despite their strengths, they are still very far from modeling entities.\footnote{Source code for our model, the training procedure and the new dataset is published on \url{https://github.com/amore-upf/analysis-entity-centric-nns}.}\looseness=-1

The task is character identification on multi-party dialogue as posed in SemEval 2018 Task 4 \cite{choi2018semeval}.\footnote{\url{https://competitions.codalab.org/competitions/17310}.}
Models are given dialogues from the TV show \emph{Friends} and asked to link entity mentions (nominal expressions like \textit{I}, \textit{she} or \textit{the woman}) to the characters to which they refer in each case.
Figure~\ref{fig:example} shows an example, where the mentions \textit{Ross} and \textit{you} are linked to entity 335, mention \textit{I} to entity 183, etc.
Since the TV series revolves around a set of entities that recur over many scenes and episodes, it is a good benchmark to analyze whether entity-centric models learn and use entity representations for referential tasks.

\begin{figure}[t]
	\fbox{
		\parbox{.45\textwidth}{
			\textsc{Joey Tribbiani} (183):\\
			\begin{tabular}{@{~}r@{~}r@{~}r@{~}r@{~}r@{~}lr}
				"\ldots \textsl{see} 	& \textsl{\underline{Ross},} 
				& \textsl{because \underline{I}} 
				& \textsl{think \underline{you}} 
				& \textsl{love}  \textsl{\underline{her}} 
				& ." \\
				& 335 
				& 183 
				& 335
				& 306
				& 
	\end{tabular}}}
    \caption{Character identification: example.}
    \label{fig:example}
\end{figure}

Our contributions are three-fold:
First, we adapt two previous entity-centric models and show that they do better on lower frequency entities (a significant challenge for current data-hungry models) than a counterpart model that is not entity-centric, with the same model size.
Second, through analysis we provide insights into how they achieve these improvements, and argue that making models entity-centric fosters architectural decisions that result in good inductive biases.
Third, we create a dataset and task to evaluate the models' ability to encode entity information such as gender, and show that models fail at it.
More generally, our paper underscores the need for the analysis of model behavior, not only through ablation studies, but also through the targeted probing of model representations~\cite{linzen2016assessing,conneau+18}.\looseness=-1


\section{Related Work}
\label{sec:related}

\paragraph{Modeling.} Various memory architectures have been proposed that are not specifically for entity-centric models, but could in principle be employed in them~\cite{graves2014neural,sukhbaatar2015end,Joulin2015,bansal2017relnet}.
The two models we base our results on \cite{henaff16:entitynetworks,aina2018semeval} were explicitly motivated as entity-centric.
We show that our adaptations yield good results and provide a closer analysis of their behavior.

\paragraph{Tasks.}
The task of entity linking has been formalized as resolving entity mentions to referential entity entries in a knowledge repository, mostly Wikipedia 
 (\citeauthor{bunescu2006using},~\citeyear{bunescu2006using}; \citeauthor{mihalcea2007wikify},~\citeyear{mihalcea2007wikify} and much subsequent work; for recent approaches see \citeauthor{francis2016capturing},~\citeyear{francis2016capturing}; \citeauthor{chen2018bilinear},~\citeyear{chen2018bilinear}).
In the present entity linking task, only a list of entities is given, without associated encyclopedic entries, and information about the entities needs to be acquired from scratch through the task; note the analogy to how a human audience might get familiar with the TV show characters by watching it.
Moreover, it addresses multiparty dialogue (as opposed to, typically, narrative text), where speaker information is crucial.
A task closely related to entity linking is \emph{coreference resolution}, i.e., predicting which portions of a text refer to the same entity (e.g., \textit{Marie Curie} and \textit{the scientist}).
This typically requires clustering mentions that refer to the same entity \cite{pradhan2011conll}.
Mention clusters essentially correspond to entities, and recent work on coreference and language modeling has started exploiting an explicit notion of entity \cite{haghighi2010coreference,clark2016improving,yang2016reference}.
Previous work both on entity linking and on coreference resolution (cited above, as well as \citeauthor{wiseman2016learning},~\citeyear{wiseman2016learning}) often presents more complex models that incorporate e.g.\ hand-engineered features.
In contrast, we keep our underlying model basic since we want to systematically analyze how certain architectural decisions affect performance.
For the same reason we deviate from previous work to entity linking that uses a specialized coreference resolution module (e.g., \citeauthor{ChenZhouChoi:17},~\citeyear{ChenZhouChoi:17}).

\paragraph{Analysis of Neural Network Models.}
Our work joins a recent strand in NLP that systematically analyzes what different neural network models learn about language \cite[a.o.]{linzen2016assessing,kadar2017representation,conneau+18,gulordava2018colorless,nematzadeh2018evaluating}.
This work, like ours, has yielded both positive and negative results: 
There is evidence that they learn complex linguistic phenomena of morphological and syntactic nature, like long distance agreement \cite{gulordava2018colorless,giulianelli2018under}, but less evidence that they learn how language relates to situations; for instance, \citet{nematzadeh2018evaluating} show that memory-augmented neural models fail on tasks that require keeping track of inconsistent states of the world.

\section{Models}
\label{sec:all_models}
We approach character identification as a classification task, and compare a baseline LSTM \cite{hochreiter1997long} with two models that enrich the LSTM with a memory module designed to learn and use entity representations.
LSTMs are the workhorse for text processing, and thus a good baseline to assess the contribution of this module.
The LSTM processes text of dialogue scenes one token at a time, and the output is a probability distribution over the entities (the set of entity IDs are given).\looseness=-1

\subsection{Baseline: \textsc{biLSTM}}

\begin{figure}[t] 
	\centering
	\includegraphics[width=.9\columnwidth]{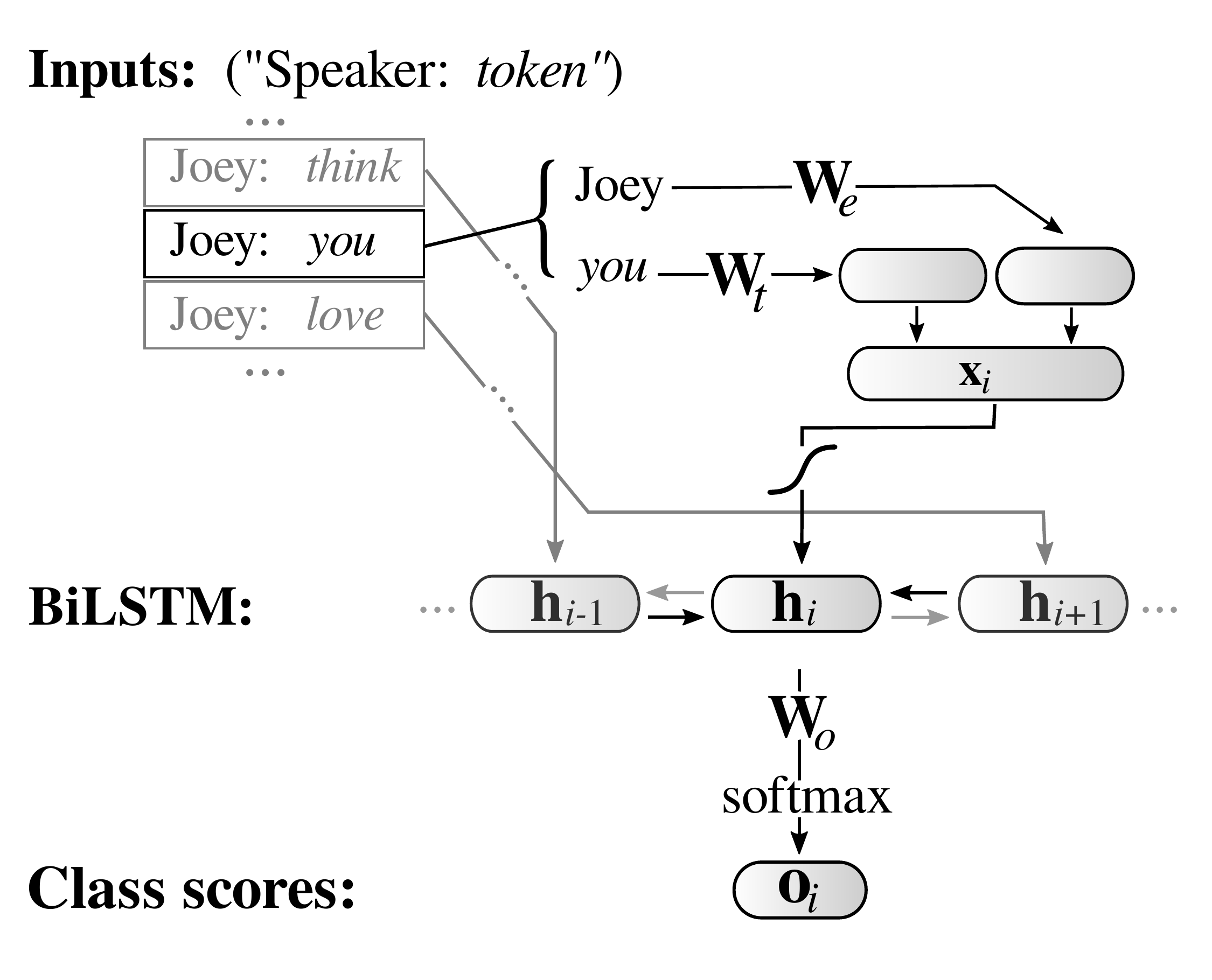}
	\caption{\textsc{biLSTM} applied to ``...think you love...'' as spoken by Joey (from Figure~\ref{fig:example}), outputting class scores for mention ``you'' (bias~$\mathrm{b}_o$ not depicted).}
	\label{fig:noentlib}
\end{figure}

The \textsc{biLSTM} model is depicted in Figure~\ref{fig:noentlib}.
It is a standard bidirectional LSTM \cite{graves2005bidirectional}, with the difference with most uses of LSTMs in NLP that we incorporate speaker information in addition to the linguistic content of utterances.

The model is given chunks of dialogue (see Appendix for hyperparameter settings such as the chunk size).
At each time step $i$, one-hot vectors for token $\mathbf{t}_i$ and speaker entities $\mathbf{s}_i$ are embedded via two distinct matrices $\mathbf{W}_t$ and $\mathbf{W}_e$ and concatenated to form a vector~$\mathbf{x}_i$ (Eq.~\ref{eq:input}, where $\|$ denotes concatenation; note that in case of multiple simultaneous speakers ${S}_i$ their embeddings are summed).
\begin{equation}
\label{eq:input} \mathbf{x}_i = \mathbf{W}_t \ \mathbf{t}_i \ \| \displaystyle \sum_{\mathbf{s} \in S_i} \mathbf{W}_e \ \mathbf{s}
\end{equation}
The vector $\mathbf{x}_i$ is fed through the nonlinear activation function $\text{tanh}$ and input to a bidirectional LSTM.
The hidden state $\overrightarrow{\mathbf{h}}_{\! i}$ of a \emph{uni}directional LSTM for the $i$\textsuperscript{th} input is recursively defined as a combination of that input with the LSTM's previous hidden state $\overrightarrow{\mathbf{h}}_{\! i-1}$.
For a \emph{bi}directional LSTM, the hidden state $\mathbf{h}_i$ is the concatenation of the hidden states $\overrightarrow{\mathbf{h}}_{\! i}$ and $\overleftarrow{\mathbf{h}}_{\! i}$ of two unidirectional LSTMs which process the data in opposite directions (Eqs.~\ref{eq:hidden}-\ref{eq:hidden3}).\looseness=-1
\begin{gather}
\label{eq:hidden} \overrightarrow{\mathbf{h}_i} = \text{LSTM}(\text{tanh}(\mathbf{x}_i), \overrightarrow{\mathbf{h}}_{\! i-1}) \\
\label{eq:hidden2} \overleftarrow{\mathbf{h}_i} = \text{LSTM}(\text{tanh}(\mathbf{x}_i), \overleftarrow{\mathbf{h}}_{\! i+1}) \\
\label{eq:hidden3} \mathbf{h}_i = \overrightarrow{\mathbf{h}_i} \ \| \ \overleftarrow{\mathbf{h}_i}
\end{gather}
For every entity mention $\mathbf{t}_i$ (i.e., every token\footnote{For multi-word mentions this is done only for the last token in the mention.} that is tagged as referring to an entity), we obtain a distribution over all entities,~\mbox{$\mathbf{o}_i \in [0,1]^{1 \times N}$}, by applying a linear transformation to its hidden state~$\mathbf{h}_i$ (Eq.~\ref{eq:output2a}), and feeding the resulting~$\mathbf{g}_i$ to a softmax classifier~(Eq.~\ref{eq:output2}).
\begin{gather}
\label{eq:output2a} \mathbf{g}_i = \mathbf{W}_o \ \mathbf{h}_i + \mathbf{b}_o \\
\label{eq:output2}\mathbf{o}_i =  \text{softmax} (\mathbf{g}_i)
\end{gather}
Eq.~\ref{eq:output2a} is where the other models will diverge. 

\subsection{\textsc{EntLib} (Static Memory)}

\begin{figure}[t] 
	\centering
	\includegraphics[width=.9\columnwidth]{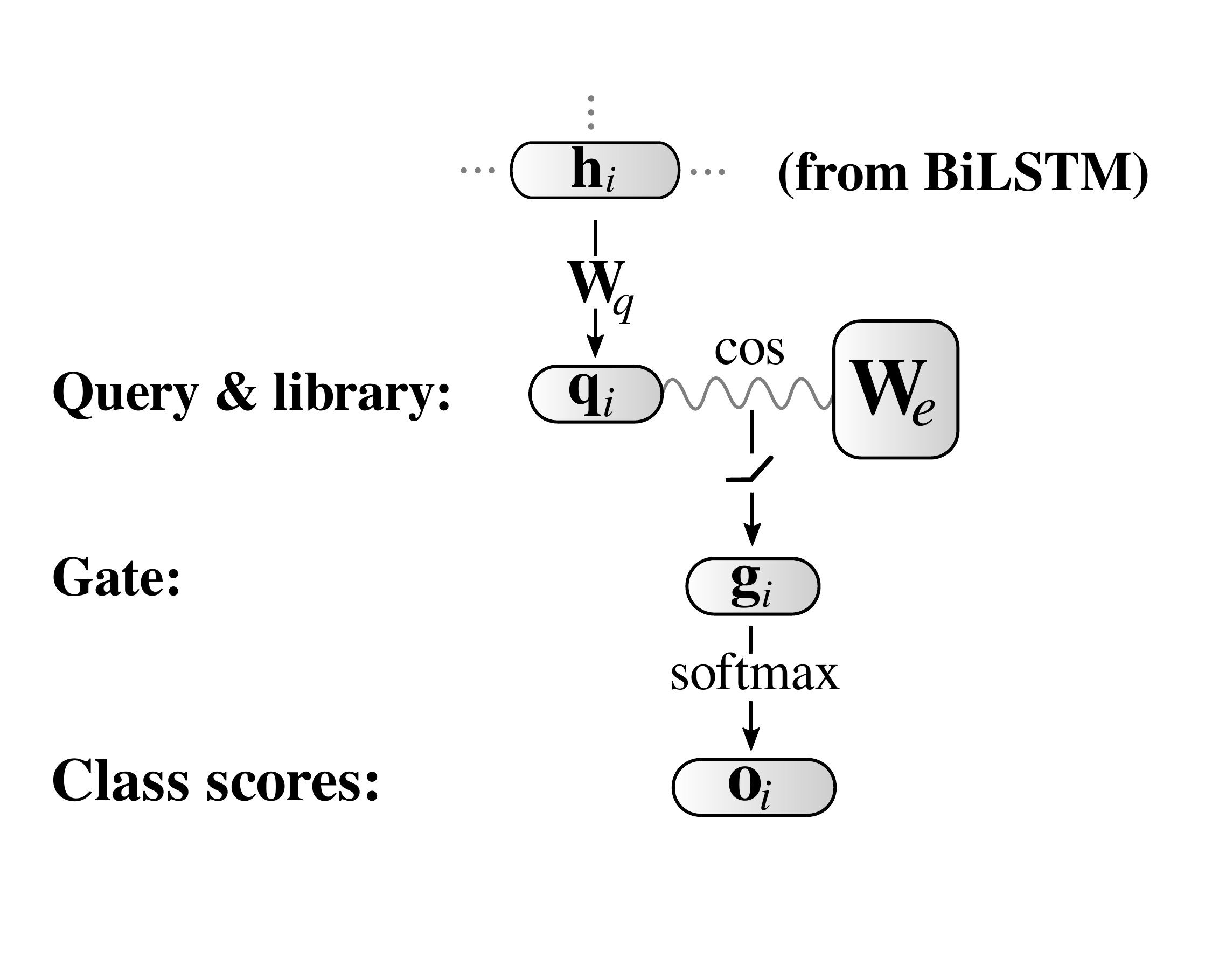}
	\caption{\textsc{EntLib}; everything before~$\mathrm{h_i}$, omitted here, is the same as in Figure~\ref{fig:noentlib}.}
	\label{fig:static}
\end{figure}
The \textsc{EntLib} model (Figure~\ref{fig:static}) is an adaptation of our previous work in \citet{aina2018semeval}, which was the winner of the SemEval 2018 Task 4 competition.
This model adds a simple memory module that is expected to represent entities because its vectors are tied to the output classes (accordingly, \citeauthor{aina2018semeval},~\citeyear{aina2018semeval},~call this module \textit{entity library}).
We call this memory `static', since it is updated only during training, after which it remains fixed.

Where \textsc{biLSTM} maps the hidden state $\mathbf{h}_i$ to class scores $\mathbf{o}_i$ with a single transformation (plus softmax), \textsc{EntLib} instead takes two steps: It first transforms $\mathbf{h}_i$ into a `query' vector $\mathbf{q}_i$ (Eq.~\ref{eq:query}) that it will then use to query the entity library.
As we will see, this mechanism helps dividing the labor between representing the context (hidden layer) and doing the prediction task (query layer).
\begin{equation}
	\label{eq:query} \mathbf{q}_i = \mathbf{W}_q \ \mathbf{h}_i + \mathbf{b_q}
\end{equation}
A weight matrix $\mathbf{W}_e$ is used as the entity library, which is the same as the speaker embedding in Eq.~\ref{eq:input}: the query vector $\mathbf{q}_i \in \mathbb{R}^{1 \times k}$ is compared to each vector in $\mathbf{W}_e$ (cosine), and a \emph{gate} vector $\mathbf{g}_i$ is obtained by applying the $\text{ReLU}$ function to the cosine similarity scores (Eq.~\ref{eq:gate_static}).%
\footnote{In \newcite{aina2018semeval}, the gate did not include the ReLU nonlinear activation function. Adding it improved results.}
Thus, the query extracted from the LSTM's hidden state is used as a soft pointer over the model's representation of the entities.
\begin{equation}
\label{eq:gate_static} \mathbf{g}_i = \text{ReLU} ( \text{cos} (\mathbf{W}_e, \mathbf{q}_i) )
\end{equation}
As before, a softmax over $\mathbf{g}_i$ then yields the distribution over entities (Eq.~\ref{eq:output2}).
So, in the \textsc{EntLib} model Eqs.~\ref{eq:query} and~\ref{eq:gate_static} together take the place of Eq.~\ref{eq:output2a} in the \textsc{biLSTM} model.

Our implementation differs from \mbox{\newcite{aina2018semeval}} in one important point that we will show to be relevant to model less frequent entities (training also differs, see Section~\ref{sec:friends-task}): The original model did not do parameter sharing between speakers and referents, but used two distinct weight matrices.

Note that the contents of the entity library in \textsc{EntLib} do not change during forward propagation of activations, but only during backpropagation of errors, i.e., during training, when the weights of $\mathbf{W}_e$ are updated.
If anything, they will encode permanent properties of entities, not properties that change within a scene or between scenes or episodes, which should be useful for reference.
The next model attempts to overcome this limitation.

\subsection{\textsc{EntNet} (Dynamic Memory)}
\begin{figure}[t] 
	\centering
	\includegraphics[width=.9\columnwidth]{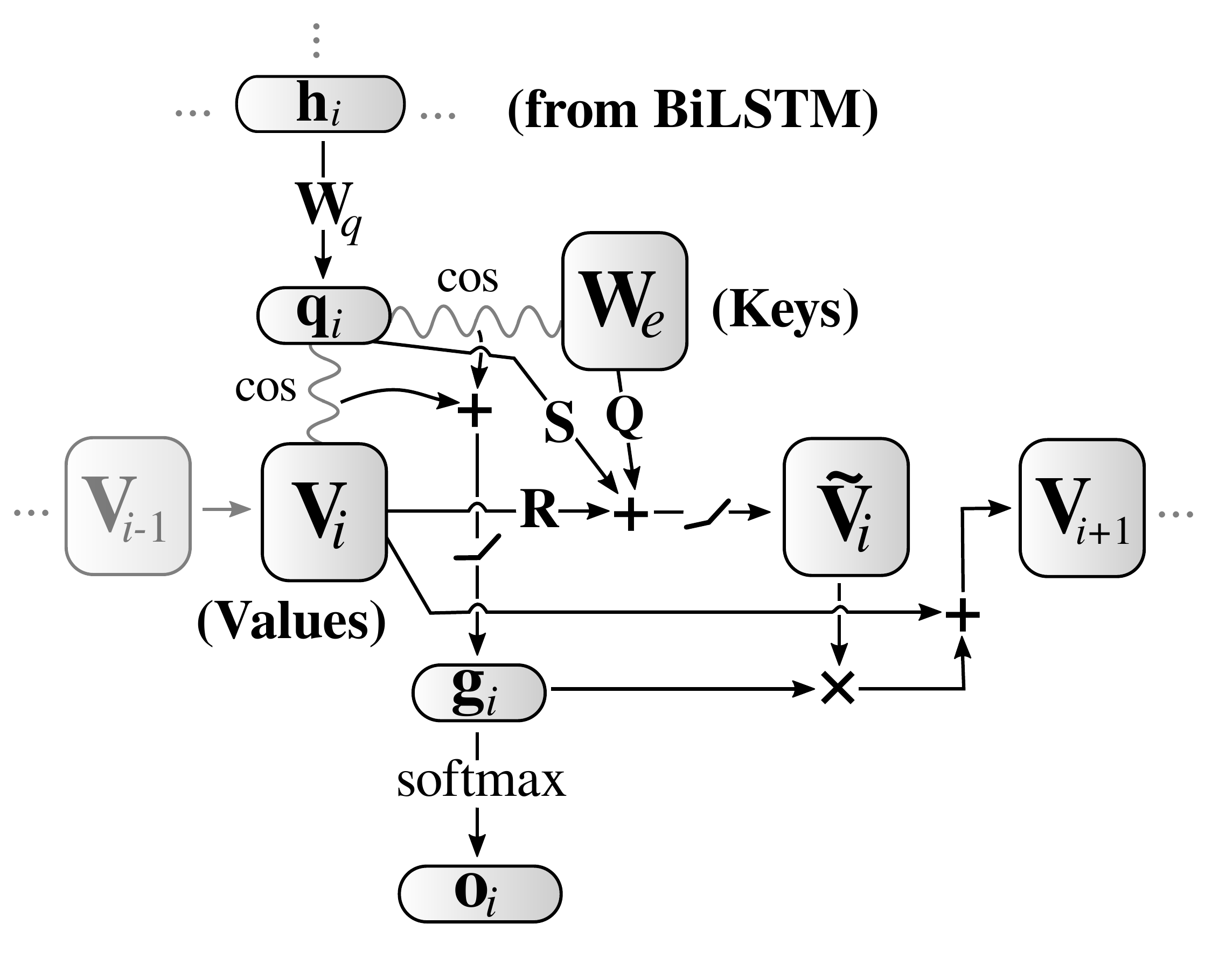}
	\caption{\textsc{EntNet}; everything before~$\mathrm{h_i}$, omitted here, is the same as in Figure~\ref{fig:noentlib}.}
	\label{fig:dynamic}
\end{figure}
\textsc{EntNet} is an adaptation of \emph{Recurrent Entity Networks} \cite[Figure~\ref{fig:dynamic}]{henaff16:entitynetworks} to the task.
Instead of representing each entity by a single vector, as in \textsc{EntLib}, here each entity is represented jointly by a context-invariant or `static' \emph{key} and a context-dependent or `dynamic' \emph{value}. 
For the keys the entity embedding $\mathbf{W}_e$ is used, just like the entity library of \textsc{EntLib}. 
But the values $\mathbf{V}_i$ can be dynamically updated throughout a scene.

As before, an entity query $\mathbf{q}_i$ is first obtained from the \textsc{biLSTM} (Eq.~\ref{eq:query}). 
Then, \textsc{EntNet} computes gate values $\mathbf{g}_i$ by estimating the query's similarity to both keys and values, as in Eq.~\ref{eq:gate_dynamic} (replacing Eq.~\ref{eq:gate_static} of \textsc{EntLib}).\footnote{Two small changes with respect to the original model (motivated by empirical results in the hyperparameter search) are that we compute the gate using cosine similarity instead of dot product, and the obtained similarities are fed through a ReLU nonlinearity instead of sigmoid.}
Output scores $\mathbf{o}_i$ are computed as in the previous models (Eq.~\ref{eq:output2}).
\begin{equation}
\label{eq:gate_dynamic} \mathbf{g}_i = \text{ReLU}(\text{cos}(\mathbf{W}_e, \mathbf{q}_i) + \text{cos}(\mathbf{V}_i, \mathbf{q}_i))
\end{equation}
The values $\mathbf{V}_i$ are initialized at the start of every scene ($i = 0$) as being identical to the keys ($\mathbf{V}_{0} = \mathbf{W}_e$).
After processing the $i^{\text{th}}$ token, new information can be added to the values.
Eq.~\ref{eq:tildev} computes this new information $\tilde{\mathbf{V}}_{i,j}$, for the $j^{\text{th}}$ entity,
where $\mathbf{Q}$, $\mathbf{R}$ and $\mathbf{S}$ are learned linear transformations and PReLU denotes the parameterized rectified linear unit \cite{PreLU}:
\begin{gather}
\label{eq:tildev}
\tilde{\mathbf{V}}_{i,j} = \text{PReLU}(\mathbf{Q}\mathbf{W}_{ej} + \mathbf{R}\mathbf{V}_{i,j} + \mathbf{S}\mathbf{q}_{i})
\end{gather}
This information $\tilde{\mathbf{V}}_{i,j}$, multiplied by the respective gate $\mathbf{g}_{i,j}$, is added to the values to be used when processing the next ($i + 1^{\text{th}}$) token (Eq.~\ref{eq:some}), and the result is normalized (Eq.~\ref{eq:norm}):
\begin{gather}
\label{eq:some}
\mathbf{V}_{i+1,j} = \mathbf{V}_{j} + \mathbf{g}_{i,j} * \tilde{\mathbf{V}}_{i,j}		
\\
\label{eq:norm}
\mathbf{V}_{i+1,j} = \frac{\mathbf{V}_{i+1,j}}{\|\mathbf{V}_{i+1,j}\|}
\end{gather}

Our adaptation of the Recurrent Entity Network involves two changes.
First, we use a biLSTM to process the linguistic utterance, while \newcite{henaff16:entitynetworks} used a simple multiplicative mask (we have natural dialogue, while their main evaluation was on bAbI, a synthetic dataset).
Second, in the original model the gates were used to retrieve and output information about the query, whereas we use them directly as output scores because our task is referential.
This also allows us to tie the keys to the characters of the Friends series as in the previous model, and thus have them represent entities (in the original model, the keys represented entity types, not instances).


\section{Character Identification}
\label{sec:friends-task}

The training and test data for the task span the first two seasons of \textsl{Friends}, divided into scenes and episodes, which were in turn divided into utterances (and tokens) annotated with speaker identity.%
\footnote{The dataset also includes automatic linguistic annotations, e.g., PoS tags, which  our models do not use.}
The set of all possible entities to refer to is given, as well as the set of mentions to resolve.
Only the dialogues and speaker information are available (e.g., no video or descriptive text). 
Indeed, one of the most interesting aspects of the SemEval data is the fact that it is dialogue (even if scripted), which allows us to explore the role of speaker information, one of the aspects of the extralinguistic context of utterance that is crucial for reference.
\savingspace{\begin{table}[t]
	\centering
	\begin{tabular}{|lrr|}
		\hline
		& Train & Test\\
		\hline
		Entities & 372 & 106\\
		Mentions & 13,280 & 2,429\\
		Scenes   & 374 & 74\\
		Episodes &  47 & 40\\
		\hline
	\end{tabular}
	\caption{Summary statistics of the SemEval 2018 Task 4 dataset. The union of entities in the training and test sets is 401.}
	\label{tab:semeval-data}
\end{table}}
We additionally used the publicly available 300-dimensional word vectors that were pre-trained on a Google News corpus with the word2vec Skip-gram model \cite{mikolov2013distributed} to represent the input tokens. 
Entity (speaker/referent) embeddings were randomly initialized.%

We train the models with backpropagation, using the standard negative log-likelihood loss function.
For each of the three model architectures we performed a random search ($>1500$ models) over the hyperparameters using cross-validation (see Appendix for details), and report the results of the best settings after retraining without cross-validation. 
The findings we report are representative of the model populations. 

\paragraph{Results.} 
We follow the evaluation defined in the SemEval task. 
Metrics are macro-average \fscore (which computes the \fscore for each entity separately and then averages these over all entities) and accuracy, in two conditions:
\emph{All entities}, with 78 classes (77 for entities that are mentioned in both training and test set of the SemEval Task, and one grouping all others), and  \emph{main entities}, with 7 classes (6 for the main characters and one for all the others).
Macro-average \fscore on all entities, the most stringent, was the criterion to define the leaderboard.
\begin{table}[t]
	\centering
	\begin{tabular}{|l@{~}r|l@{~}|l@{~}|l|l|}
		\hline
		& & \multicolumn{2}{c|}{all (78)} &
		\multicolumn{2}{c|}{main (7)} \\
		models            & \#par & F$_1$ & Acc  & F$_1$ & Acc \\
		\hline \hline
		SemEv-1st & - & 41.1  & 74.7 & 79.4 & 77.2 \\
		SemEv-2nd & - & 13.5  & 68.6 & 83.4 & 82.1 \\
		\hline
		\textsc{biLSTM}   & 3.4M & 34.4  & 74.6 & \textbf{85.0} & 83.5 \\		
		\textsc{EntLib}  & 3.3M & 49.6$^*$  & \textbf{77.6}$^*$ & 84.9 & \textbf{84.2} \\
		\textsc{EntNet} & 3.4M & \textbf{52.5}$^*$  & 77.5$^*$ & 84.8 & 83.9 \\
		\hline
	\end{tabular}
	\caption{Model parameters and results on the character identification task. First block: top systems at SemEval 2018. 
Results in the second block marked with $^*$ are statistically
          significantly better than \textsc{biLSTM} at~\mbox{$p <
            0.001$} (approximate randomization tests,
          \protect\citeauthor{noreen1989computer},~\citeyear{noreen1989computer}).}
	\label{tab:main_results}
\end{table}

Table~\ref{tab:main_results} gives our results in the two evaluations, comparing the models described in Section~\ref{sec:all_models} to the best performing models in the SemEval 2018 Task 4 competition \cite{aina2018semeval,cheoneum:18semeval}.
Recall that our goal in this paper is not to optimize performance, but to understand model behavior; however, results show that these models are worth analyzing, as that they outperform the state of the art.
All models perform on a par on main entities, but entity-centric models outperform \textsc{biLSTM} by a substantial margin when all characters are to be predicted (the difference between \textsc{EntLib} and \textsc{EntNet} is not significant).

\begin{figure}[t]
	\centering
	\includegraphics[width=0.8\columnwidth]{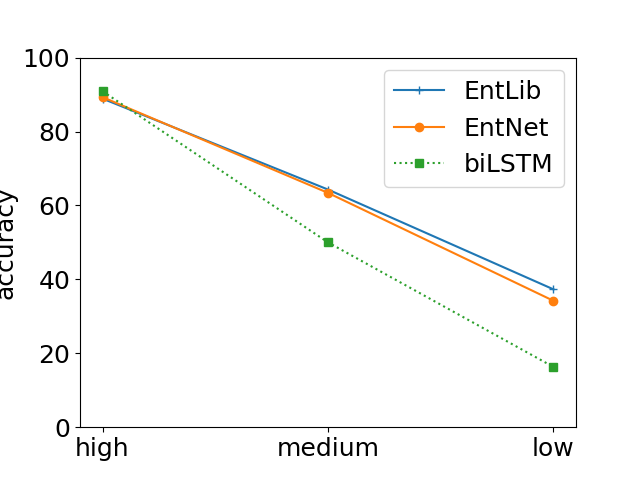}
	\caption{Accuracy on entities with high ($>$1000), medium (20--1000), and low ($<$20) frequency.
		\label{fig:acc_relative_to_frequency}}
\end{figure}

The architectures of \textsc{EntLib} and \textsc{EntNet} help with lower frequency characters, while not hurting performance on main characters.
Indeed, Figure~\ref{fig:acc_relative_to_frequency} shows that the accuracy of \textsc{biLSTM} rapidly deteriorates for less frequent entities, whereas \textsc{EntLib} and \textsc{EntNet} degrade more gracefully.
Deep learning approaches are data-hungry, and entity mentions follow the Zipfian distribution typical of language, with very few high frequency and many lower-frequency items, such that this is a welcome result.
Moreover, these improvements do not come at the cost of model complexity in terms of number of parameters, since all models have roughly the same number of parameters ($3.3-3.4$~million).\footnote{See Appendix for a computation of the models' parameters.}

\begin{figure}[t]
	\centering
	\includegraphics[width=.44\textwidth]{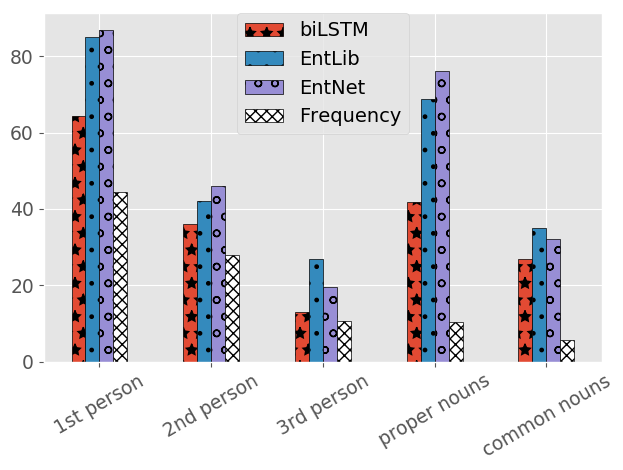}
	\caption{\fscore (\emph{all entities} condition) of the three models, per mention type, and token frequency of each mention type.\label{fig:pronouns_all2}}
\end{figure}

Given these results and the motivations for the model architectures, it would be tempting to conclude that encouraging models to learn and use entity representations helps in this referential task.
However, a closer look at the models' behavior reveals a much more nuanced picture.

Figure~\ref{fig:pronouns_all2} suggests that:
(1) models are quite good at using speaker information, as the best performance is for first person pronouns and determiners (\textit{I}, \textit{my}, etc.);
(2) instead, models do not seem to be very good at handling other contextual information or entity-specific properties, as the worst performance is for third person mentions and 
common nouns, which require both;\footnote{1st person:  \textit{I}, \textit{me}, \textit{my}, \textit{myself}, \textit{mine}; 2nd person:
  \textit{you}, \textit{your}, \textit{yourself}, \textit{yours}; 3rd person: \textit{she}, \textit{her}, \textit{herself}, \textit{hers}, \textit{he}, \textit{him}, \textit{himself}, \textit{his}, \textit{it}, \textit{itself}, \textit{its}.}
(3)~\textsc{EntLib} and \textsc{EntNet} behave quite similarly, with performance boosts in (1) and smaller but consistent improvements in (2). 
Our analyses in the next two sections confirm this picture and relate it to the models' architectures.

\section{Analysis: Architecture}

We examine how the entity-centric architectures improve over the \textsc{biLSTM} baseline on the reference task, then move to entity representations (Section~\ref{sec:ent-rep}).

\begin{figure*}[htb]
	\centering
	\begin{subfigure}{\columnwidth}
		\centering
		\includegraphics[width=\columnwidth]{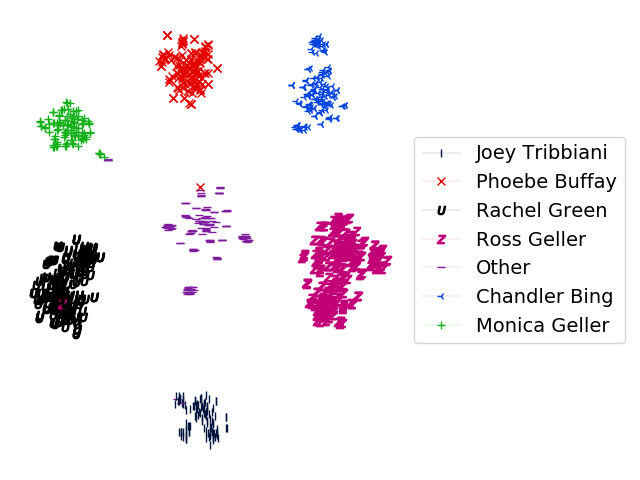}
	\end{subfigure}%
	\begin{subfigure}{\columnwidth}
		\centering
		\includegraphics[width=\columnwidth]{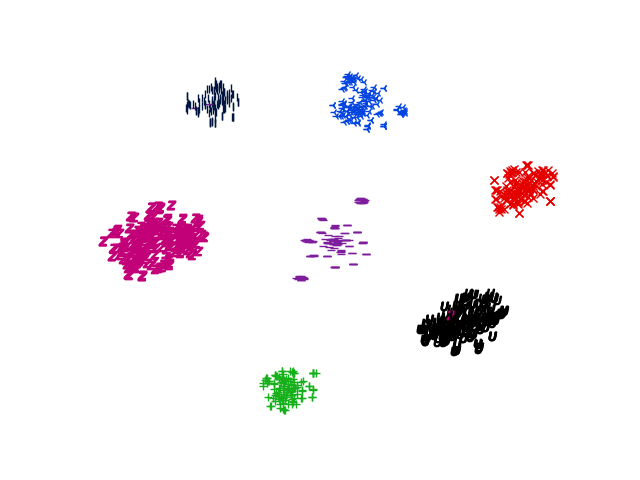}
	\end{subfigure}
	\caption{\textsc{EntLib}, 2D TSNE projections of the activations for first-person mentions in the test set, colored by the entity referred to. The mentions cluster into entities already in the hidden layer~$\mathrm{h}_i$ (left graph; query layer~$\mathrm{q}_i$ shown in the right graph). Best viewed in color.}
	\label{fig:hi-ei-1person}
\end{figure*}

\begin{figure*}[tb]
	\centering
	\begin{subfigure}{\columnwidth}		
		\centering
		\includegraphics[width=\columnwidth]{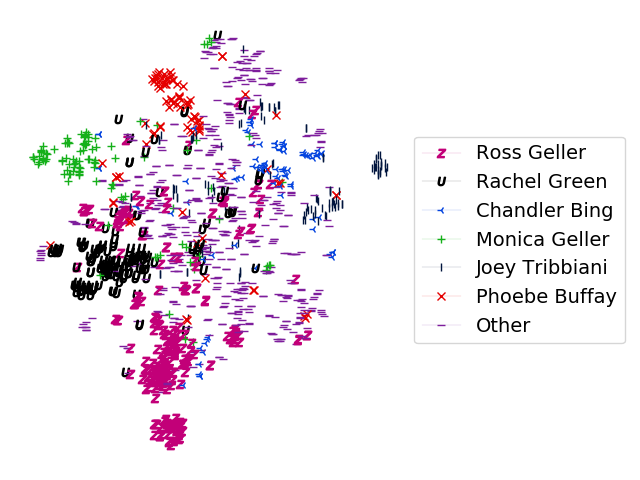}
	\end{subfigure}%
	\begin{subfigure}{\columnwidth}
		\centering	
		\includegraphics[width=\columnwidth]{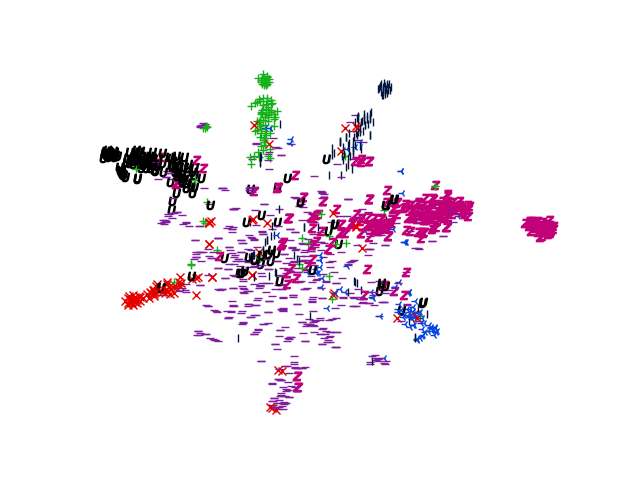}
	\end{subfigure}
	\caption{\textsc{EntLib}, 2D TSNE projections of the activations for mentions in the test set (excluding first person mentions), colored by the entity referred to. While there is already some structure in the hidden layer~$\mathrm{h}_i$ (left graph), the mentions cluster into entities much more clearly in the query~$\mathrm{q}_i$ (right graph). Best viewed in color.}
	\label{fig:hi-ei-not-1person-entlib}
\end{figure*}


\paragraph{Shared speaker/referent representation.}
We found that an important advantage of the entity-centric models, in particular for handling low-frequency entities, lies in the integrated representations they enable of entities both in their role of speakers and in their role of referents.
This explains the boost in first person pronoun and proper noun mentions, as follows.

Recall that the integrated representation is achieved by parameter sharing, using the same weight matrix $\mathbf{W}_e$ as speaker embedding and as entity library/keys.
This enables entity-centric models to learn the linguistic rule ``a first person pronoun (\textit{I, me}, etc.) refers to the speaker'' regardless of whether they have a meaningful representation of this particular entity: It is enough that speaker representations are distinct, and they are because they have been randomly initialized.
In contrast, the simple \textsc{biLSTM} baseline needs to independently learn the mapping between speaker embedding and output entities, and so it can only learn to resolve even first-person pronouns for entities for which it has enough data.

\begin{table}[tb]
	\centering
	\begin{tabular}{|lcc|}
          \hline
           model type         & main & all\\
          \hline \hline
          \textsc{biLSTM} & 0.39 & 0.02\\
          \textsc{EntLib} & 0.82 & 0.13\\
          \textsc{EntNet} & 0.92 & 0.16\\
          \hline
          \#pairs         & 21 & 22155\\
          \hline
	\end{tabular}
        \caption{RSA correlation between speaker/referent embeddings~$\mathrm{W}_e$ and token embeddings~$\mathrm{W}_t$ of the entities' names, for main entities vs.\ all entities (right)}
\label{tab:rsa-results}
\end{table}

For proper nouns (character names), entity-centric models learn to align the token embeddings with the entity representations (identical to the speaker embeddings).
We show this by using Representation Similarity Analysis (RSA) \citep{kriegeskorte2008representational}, which measures how topologically similar two different spaces are as the Spearman correlation between the pair-wise similarities of points in each space (this is necessary because entities and tokens are in different spaces).
For instance, if the two spaces are topologically similar, the relationship of entities 183 and 335 in the entity library will be analogous to the relationship between the names \emph{Joey} and \emph{Ross} in the token space.
Table~\ref{tab:rsa-results} shows the topological similarities between the two spaces, for the different model types.%
\footnote{As an entity's name we here take the proper noun that is most frequently used to refer to the entity in the training data.
Note that for the \emph{all entities} condition the absolute values are lower, but the space is much larger (over 22K pairs).
Also note that this is an instance of slow learning; models are not encoding the fact that a proper noun like \emph{Rachel} can refer to different people.}
This reveals that in entity-centric models the space of speaker/referent embeddings is topologically very similar to the space of token embeddings restricted to the entities' names, and more so than in the \textsc{biLSTM} baseline.
We hypothesize that entity-centric models can do the alignment better because referent (and hence speaker) embeddings are closer to the error signal, and thus backpropagation is more effective (this again helps with lower-frequency entities).

Further analysis revealed that in entity-centric models the beneficial effect of weight sharing between the speaker embedding and the entity representations (both $\mathbf{W}_e$) is actually restricted to first-person pronouns. For other expressions, having two distinct matrices yielded almost the same performance as having one (but still higher than the \textsc{biLSTM}, thanks to the other architectural advantage that we discuss below).

In the case of first-person pronouns, the speaker embedding given as input corresponds to the target entity. This information is already accessible in the hidden state of the LSTM. Therefore, mentions cluster into entities already at the hidden layer $h_i$, with no real difference with the query layer $q_i$ (see Figure~\ref{fig:hi-ei-1person}).

\paragraph{Advantage of query layer.}
The entity querying mechanism described above entails having an extra transformation after the hidden layer, with the query layer $\mathbf{q}$.
Part of the improved performance of entity-centric models, compared to the \textsc{biLSTM} baseline, is due not to their bias towards `entity representations' per se, but due to the presence of this extra layer.
Recall that the \textsc{biLSTM} baseline maps the LSTM's hidden state $\mathbf{h}_i$ to output scores $\mathbf{o}_i$ with a single transformation. 
\newcite{gulordava2018represent} observe in the context of Language Modeling that this creates a tension between two conflicting requirements for the LSTM: keeping track of contextual information across time steps, and encoding information useful for prediction in the current timestep.
The intermediate query layer $\mathbf{q}$ in entity-centric models alleviates this tension.
This explains the improvements in context-dependent mentions like common nouns or second and third pronouns.

\begin{table}[tb]
	\centering
	\begin{tabular}{|c|cc|cc|}
		\hline
		\textsc{biLSTM} & \multicolumn{2}{c|}{\textsc{EntLib}} & \multicolumn{2}{c|}{\textsc{EntNet}} \\
		$\mathbf{h}_i$  & $\mathbf{h}_i$ &  $\mathbf{q}_i$ & $\mathbf{h}_i$ & $\mathbf{q}_i$ \\\hline
		0.34  & 0.24 & 0.48 & 0.27 & 0.60 \\ \hline
	\end{tabular}
	\caption{\label{tab:hi-ei}Average cosine similarity of mentions with the same referent.}
\end{table}

We show this effect in two ways.
First, we compare the average mean similarity $s$ of mention pairs
~\mbox{$T_e = \{(t_k, t_{k'}) | \; t_k \rightarrow e \wedge k \neq k'\}$} referring to the same entity~$e$ in the hidden layer (Eq.~\ref{eq:mention_pairs}) and the query layer.%
\footnote{For the query layer, Eq.~\ref{eq:mention_pairs} is equivalent, with $\text{cos}(q_{t_k}, q_{t_{k'}})$.}
\begin{equation}
\label{eq:mention_pairs}
s = \frac{1}{|E|}\sum_{e \in E} \frac{1}{|T_e|} \sum_{(t_k,t_{k'}) \in T_e}  \text{cos}(h_{t_k}, h_{t_{k'}})
\end{equation}
Table~\ref{tab:hi-ei} shows that, in entity-centric models, this similarity is lower in the hidden layer $\mathbf{h}_i$ than in the case of the \textsc{biLSTM} baseline, but in the query layer $\mathbf{q}_i$ it is instead much higher.
The hidden layer thus is representing other information than referent-specific knowledge, and the query layer can be seen as extracting referent-specific information from the hidden layer.
Figure~\ref{fig:hi-ei-not-1person-entlib} visually illustrates the division of labor between the hidden and query layers.
Second, we compared the models to variants where the cosine-similarity comparison is replaced by an ordinary dot-product transformation, which converts the querying mechanism into a simple further layer.
These variants performed almost as well on the reference task, albeit with a slight but consistent edge for the models using cosine similarity.

\paragraph{No dynamic updates in \textsc{EntNet}.}
A surprising negative finding is that \textsc{EntNet} is not using its dynamic potential on the referential task.
We confirmed this in two ways. 
First, we tracked the values $\mathbf{V}_i$ of the entity representations and found that the pointwise difference in $\mathbf{V}_i$ at any two adjacent time steps $i$ tended to zero.
Second, we simply switched off the update mechanism during testing and did not observe any score decrease on the reference task.
\textsc{EntNet} is thus only using the part of the entity memory that it shares with \textsc{EntLib}, i.e., the keys $\mathbf{W_e}$, which explains their similar performance.

This finding is markedly different from \citet{henaff16:entitynetworks}, where for instance the BaBI tasks could be solved only by dynamically updating the entity representations.
This may reflect our different language modules: since our LSTM module already has a form of dynamic memory, unlike the simpler sentence processing module in \citet{henaff16:entitynetworks}, it may be that the LSTM takes this burden off of the entity module.
An alternative is that it is due to differences in the datasets.
We leave an empirical comparison of these potential explanations for future work, and focus in Section~\ref{sec:ent-rep} on the static entity representations $\mathbf{W}_e$ that \textsc{EntNet} essentially shares with \textsc{EntLib}.\looseness=-1


\section{Analysis: Entity Representations}
\label{sec:ent-rep}
\begin{figure}[tb]
	\fbox{\parbox{0.9\columnwidth}{
			This \underline{person} is \{a/an/the\} \textsc{\textless property\textgreater} [and \{a/an/the\} \textsc{\textless property\textgreater}]\{0,2\}.
			
			\vspace{0.1cm}
			\textit{This person is the brother of Monica Geller.}\\
			\textit{This person is a paleontologist and a man.}
	}}
	\caption{Patterns and examples (in italics) of the dataset for information extraction as entity linking.}
	\label{fig:example2}
\end{figure}
The foregoing demonstrates that entity-centric architectures help in a reference task, but not that the induced representations in fact contain meaningful entity information. 
In this section we deploy these representations on a new dataset, showing that they do not---not even for basic information about entities such as gender.

\paragraph{Method.} We evaluate entity representations with an information extraction task including attributes and relations, using information from an independent, unstructured knowledge base---the  Friends Central Wikia.\footnote{\url{http://friends.wikia.com}.}
To be able to use the models as is, we set up the task in terms of entity linking, asking models to solve the reference of natural language descriptions that uniquely identify an entity.
For instance, given \emph{This person is the brother of Monica Geller.}, the task is to determine that \emph{person} refers to \textsl{Ross Geller}, based on the information in the sentence.%
\footnote{The referring expression is the whole DP, \emph{This person}, but we follow the method in \citealt{aina2018semeval} of asking for reference resolution at the head noun.}
The information in the descriptions was in turn extracted from the Wikia.
We do not retrain the models for this task in any way---we simply deploy them.

We linked the entities from the Friends dataset used above to the Wikia through a semi-automatic procedure that yielded 93 entities, and parsed the Wikia to extract their attributes (\textsl{gender} and \textsl{job}) and relations (e.g.,~\textsl{sister}, \textsl{mother-in-law}; see Appendix for details). 
We automatically generate the natural language descriptions with a simple pattern (Figure~\ref{fig:example2}) from combinations of properties that uniquely identify a given entity within the set of Friends characters.\footnote{Models require inputting a speaker; we use speaker \textsc{Unknown}.}
We consider unique descriptions comprising at most 3 properties.
Each property is expressed by a noun phrase, whereas the article is adapted (definite or indefinite) depending on whether that property applies to one or several entities in our data. 
This yields 231 unique natural language descriptions of 66 characters, created  on the basis of overall $61$~relation types and $56$~attribute values.\looseness=-1

 \begin{table}[tb]
   \centering
 		\begin{tabular}{|l|c|cc|}
 			\hline 
                   model &  description &  gender &  job\\ 
 			\hline \hline
 			\textsc{Random}          
 			& 1.5
 			& 50
 			& 20 \\
 			\textsc{biLSTM}
 			& 0.4
 			& -
 			& - \\		
 			\textsc{EntLib}	
 			& 2.2
 			& 55
 			& 27 \\		
 			\textsc{EntNet}	
 			& 1.3
 			& 61
 			& 24 \\
 			\hline
 		\end{tabular}
 	\caption{\label{tab:entities}\label{tab:results-att-rel-prediction} Results on the attribute and relation prediction task: percentage accuracy for natural language descriptions, mean reciprocal rank of characters for single attributes (lower is worse).}
\end{table}

\paragraph{Results.}
The results of this experiment are negative:
The first column of Table~\ref{tab:results-att-rel-prediction} shows that models get accuracies near 0.

A possibility is that models do encode information in the entity representations, but it doesn't get used in this task because of how the utterance is encoded in the hidden layer, or that results are due to some quirk in the specific setup of the task.
However, we replicated the results in a setup that does not encode whole utterances but works with single attributes and relations.
While the methodological details are in the Appendix, the `gender' and `job' columns of Table~\ref{tab:results-att-rel-prediction} show that results are a bit better in this case but models still perform quite poorly:
Even in the case of an attribute like gender, which is crucial for the resolution of third person pronouns (\textit{he/she}), the models' results are quite close to that of a random baseline.

Thus, we take it to be a robust result that entity-centric models trained on the SemEval data do not learn or use entity information---at least as recoverable from language cues.
This, together with the remainder of the results in the paper, suggests that models rely crucially on speaker information, but hardly on information from the linguistic context.\footnote{Note that 44\% of the mentions in the dataset are first person, for which linguistic context is irrelevant and the models only need to recover the relevant speaker embedding to succeed.
However, downsampling first person mentions did not improve results on the other mention types.}
Future work should explore alternatives such as pre-training with a language modeling task, which could improve the use of context.

\section{Conclusions}
Recall that the motivation for entity-centric models is the hypothesis that incorporating entity representations into the model will help it better model the language we use to talk about them.
We still think that this hypothesis is plausible.
However, the architectures tested do not yet provide convincing support for it, at least for the data analyzed in this paper.

On the positive side, we have shown that framing models from an entity-centric perspective makes it very natural to adopt architectural decisions that are good inductive biases.
In particular, by exploiting the fact that both speakers and referents are entities, these models can do more with the same model size, improving results on less frequent entities and emulating rule-based behavior such as ``a first person expression refers to the speaker''.
On the negative side, we have also shown that they do not yield operational entity representations, and that they are not making good use of contextual information for the referential task.

More generally, our paper underscores the need for model analysis to test whether the motivations for particular architectures are borne out in how the model actually behaves when it is deployed.

\section*{Acknowledgments}
We gratefully acknowledge Kristina Gulordava and Marco Baroni for the feedback, advice and support. We are also grateful to the anonymous reviewers for their valuable comments. This project has received funding from the European Research Council (ERC) under the European Union’s Horizon 2020 research and innovation programme (grant agreement No 715154), and from the Spanish Ram\'on y Cajal programme (grant RYC-2015-18907). We are grateful to the NVIDIA Corporation for the donation of GPUs used for this research. We are also very grateful to the Pytorch developers. This paper reflects the authors' view only, and the EU is not responsible for any use that may be made of the information it contains.
\begin{flushright}
\includegraphics[width=0.8cm]{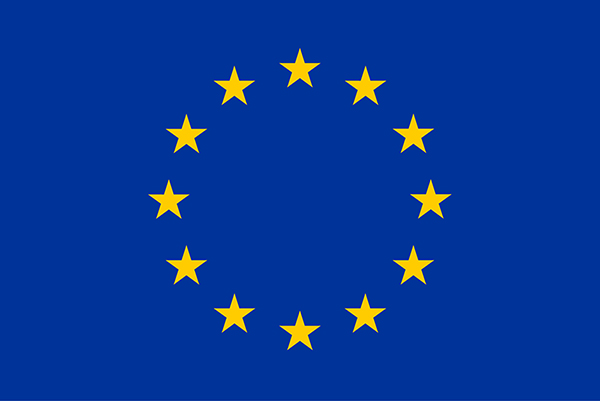}  
\includegraphics[width=0.8cm]{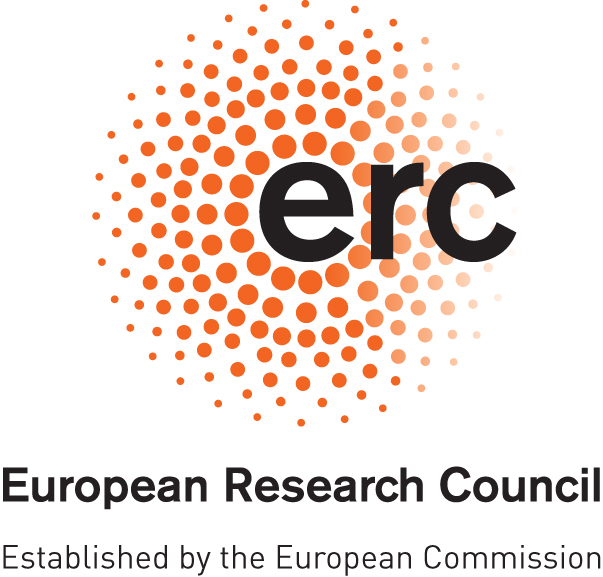} 
\end{flushright}

\bibliography{emnlp2018-short}
\bibliographystyle{acl_natbib}

\appendix
\section{Appendices}
\label{sec:appendix}

\subsection{Hyperparameter search}
Besides the LSTM parameters, we optimize the token embeddings $\mathbf{W}_t$, the entity/speaker embeddings $\mathbf{W}_e$, as well as $\mathbf{W}_o$, $\mathbf{W}_q$, and their corresponding biases, where applicable (see Section~\ref{sec:all_models}).
We used five-fold cross-validation with early stopping based on the validation score.
We found that most hyperparameters could be safely fixed the same way for all three types.
Specifically, our final models were all trained in batch mode using the Adam optimizer \cite{kingma2017adam},
with each batch covering 25 scenes given to the model in chunks of 750 tokens.
The token embeddings ($\mathbf{W}_t$) are initialized with the $300$-dimensional word2vec vectors, $\mathbf{h}_i$ is set to $500$~units, and entity (or speaker) embeddings ($\mathbf{W}_e$) to $k~\mbox{= 150}$~units.%
With this hyperparameter setting, \textsc{EntLib} has fewer parameters than \textsc{biLSTM}: the linear map $\mathbf{W}_o$ of the latter ($500 \times 401$) is replaced by the query extractor $\mathbf{W}_q$ ($500 \times 150$) followed by (non-parameterized) similarity computations.
	This holds even if we take into account that the entity embedding $\mathbf{W}_e$ used in both models contains 274 entities that are never speakers and that are, hence, used by \textsc{EntLib} but not by \textsc{biLSTM}.
	
Our search also considered different types of activation functions in different places, with the architecture presented above, i.e.,~tanh before the LSTM and ReLU in the gate, robustly yielding the best results.
Other settings tested---randomly initialized token embeddings, self-attention on the input layer, and a uni-directional LSTM---did not improve performance.

We then performed another random search ($>200$ models) over the remaining hyperparameters:
learning rate (sampled from the logarithmic interval $0.001$--$0.05$), dropout before and after LSTM (sampled from $0.0$--$0.3$ and $0.0$--$0.1$, respectively), weight decay (sampled from $10^{-6}$--$10^{-2}$) and penalization, i.e., whether to decrease the relative impact of frequent entities by dividing the loss for an entity by the square root of its frequency.
This paper reports the best model of each type, i.e., \textsc{biLSTM}, \textsc{EntLib}, and \textsc{EntNet},
after training on all the training data without cross-validation for 20, 80 and 80 epochs respectively (numbers selected based on tendencies in training histories).
These models had the following parameters:\\\\
\begin{tabular}{l|lll}
	& \textsc{biLSTM}\!\!\! & \textsc{EntLib}\!\!\! & \textsc{EntNet} \\\hline
	learning rate: & 0.0080 & 0.0011 & 0.0014 \\
	dropout pre & 0.2 & 0.2 & 0.0 \\
	dropout post: & 0.0 & 0.02 & 0.08 \\
	weight decay:\! & 1.8e-6 & 4.3e-6 &  1.0e-5 \\
	penalization: & no & yes & yes
\end{tabular}
\section{Attribute and relation extraction}

 \begin{table*}[htb]
 	\begin{center}
 		\begin{tabular}{|l|cccc|}
 			\hline \bf Model 
 			& \multicolumn{2}{c}{\textbf{Gender} (93;2)} 
 			& \multicolumn{1}{c}{\bf Occupation} 
 			& \multicolumn{1}{c|}{\bf Relatives} \\
 			& \multicolumn{1}{c}{(wo)man} 
 			& \multicolumn{1}{c}{(s)he} 
 			& \multicolumn{1}{c}{(24;17)} 
 			& \multicolumn{1}{c|}{(56;24)} \\
 			\hline \hline
 			\textsc{Random}          
 			& .50 & .50 
 			& .20 & .16 \\
 			\textsc{EntLib}	
 			& .55 & .58 
 			& .27 & .22\\		
 			\textsc{EntNet}	
 			& \textbf{.61} & .56 
 			& .24 & \textbf{.26}\\
 			
 			\hline
 		\end{tabular}
 	\end{center}
 	\caption{\label{tab:entities-app} Results on the attribute prediction task (mean reciprocal rank; from 0 (worst) to 1 (best)). The number of considered test items and candidate values, respectively, are given in the parentheses. For gender, we used \textit{(wo)man} and \textit{(s)he} as word cues for the values (fe)male.}
 \end{table*}

\subsection{Details of the dataset}
We performed a two-step procedure to extract all the available data for the SemEval characters. 
First, using simple word overlap, we automatically mapped the $401$~SemEval names to the characters in the database. 
In a second, manual step, we corrected these mappings  and added links that were not found automatically due to name alternatives, ambiguities or misspellings (e.g.,~SemEval \textsl{Dana} was mapped to \textsl{Dana Keystone}, and \textsl{Janitor} to \textsl{The Zoo Employee}). 
In total, we found~$93$ SemEval entities in Friends Central, and we extracted their attributes (gender and job) and their mutual relationships (relatives). 

\subsection{Alternative setup}
We use the same models, i.e. \textsc{EntLib} and \textsc{EntNet} trained on Experiment 1, and (without further training) extract representations for the entities from them. 
The former are directly obtained from the entity embedding~$\mathbf{W}_e$ of each model. 

In the \textbf{attribute prediction} task, we are given an attribute (e.g.,~\textsl{gender}), and all its possible values ~$\mathcal{V}$ (e.g.,~\mbox{$\mathcal{V=}$\textit{ \{woman, man \}}}). 
We formulate the task as, given a character (e.g.,~\textsl{Rachel}), producing a ranking of the possible values in descending order of their similarity to the character, where similarity is computed by measuring the cosine of the angle between their respective vector representations in the entity space. We obtain representations of attributes values, in the same space as the entities, by inputting each attribute value as a separate utterance to the models, and extracting the corresponding entity query ($\mathbf{q}_i$).
Since the models also expect a speaker for each utterance, we set it to either \textsl{all entities}, \textsl{main entities}, a \textsl{random} entity, or \textsl{no} entity (i.e.,~speaker embedding with zero in all units), and report the best results.

We evaluate the rankings produced for both tasks in terms of mean reciprocal rank \cite{craswell2009mean}, scoring from 0 to 1 (from worst to best) the position of the target labels in the ranking. 
The two first columns Table~\ref{tab:entities-app} presents the results.
Our models generally perform poorly on the tasks, though outperforming a random baseline. Even in the case of an attribute like gender, which is crucial for the resolution of third person pronouns, the models' results are still very close to that of the random baseline.

Instead, the task of \textbf{relation prediction} is to, given a pair of characters (e.g., \textsl{Ross} and \textsl{Monica}), predict the relation~$R$ which links them (e.g., \textit{sister, brother-in-law, nephew}; we found 
$24$ relations that applied to at least two pairs).
We approach this following the vector offset method introduced by \newcite{mikolov2013linguistic} for semantic relations between words. 
This leverages on regularities in the embedding space, taking the embeddings of pairs that are connected by the same relation to have analogous spatial relations. 
For two pairs of characters~$(a, b)$ and~$(c, d)$ which bear the same relation~$R$, we assume~\mbox{$\mathbf{a} - \mathbf{b} \approx \mathbf{c} - \mathbf{d}$ }to hold for their vector representations. For a target pair~\mbox{$(a, b)$} and a relation~$R$, we then compute the following measure:
\begin{equation}
\label{equ:rel-sim}
\text{s}_{rel}((a,b), R) = \frac{\sum_{(x, y) \in R} \text{cos}(\mathbf{a} - \mathbf{b}, \mathbf{x} - \mathbf{y})}{|R|}
\end{equation}
\noindent Equation~(\ref{equ:rel-sim}) computes the average relational similarity between the target character pair and the exemplars of that relation (excluding the target itself), where the relational similarity is estimated as the cosine between the vector differences of the two pairs of entity representations respectively. Due to this setup, we restrict to predicting relation types that apply to at least two pairs of entities.
For each target pair~\mbox{$(a,b)$}, we produce a rank of candidate relations in descending order of their scores~\mbox{$\text{s}_{rel}$}. 
Table~\ref{tab:entities-app} contains the results, again above baseline but clearly very poor.


\end{document}